\documentclass[]{scrartcl} 

\usepackage{hyperref}
\usepackage{scrlayer-scrpage}
\usepackage{lmodern}

\title{CAD2Real: Deep learning with domain randomization of CAD data for 3D pose estimation of electronic control unit housings}

\author{Simon B\"auerle\textsuperscript{1,2}, Jonas Barth\textsuperscript{2}, Elton Tavares de Menezes\textsuperscript{2}, \\
	Andreas Steimer\textsuperscript{2}, Ralf Mikut\textsuperscript{1} \\ \\
	\textsuperscript{1} Institute for Automation and Applied Informatics,\\
	Karlsruhe Institute of Technology, Karlsruhe, Germany \\ 
	\textsuperscript{2} Robert Bosch GmbH \\ \\
	E-Mail: \{simon.baeuerle, ralf.mikut\}@kit.edu}

\date{}

\usepackage[utf8]{inputenc} 
\usepackage[T1]{fontenc} 

\usepackage{amsmath,amssymb,amsfonts,mathtools,amsthm} 

\usepackage[babel,autostyle]{csquotes}

\usepackage{tabularx}
\newcolumntype{L}[1]{>{\raggedright\arraybackslash}p{#1}} 
\newcolumntype{C}[1]{>{\centering\arraybackslash}p{#1}} 
\newcolumntype{R}[1]{>{\raggedleft\arraybackslash}p{#1}} 
\newcolumntype{K}[1]{>{\raggedright\arraybackslash}m{#1}} 
\newcolumntype{X}[1]{>{\centering\arraybackslash}m{#1}} 
\newcolumntype{E}[1]{>{\raggedleft\arraybackslash}m{#1}} 

\usepackage{booktabs}

\usepackage{paralist}   

\usepackage[noend]{algpseudocode}

\usepackage{listings} 
\usepackage{subcaption}  

\usepackage{epstopdf}
\usepackage{xcolor}
\selectcolormodel{gray}  

\usepackage{scrhack}
\usepackage{varwidth}
\usepackage{rotating} 
\usepackage[defaultlines=2,all]{nowidow}  


\begin{document}

\hyphenpenalty=1000 
\lefthyphenmin=2 
\righthyphenmin=3 

\cleardoublepage
\setcounter{page}{1}
\pagestyle{scrheadings}

\setnowidow[2]
\setnoclub[2]
	
\maketitle

\numberwithin{equation}{section}


\section*{Abstract}
Electronic control units (ECUs) are essential for many automobile components, e.g. engine, anti-lock braking system (ABS), steering and airbags.
For some products, the 3D pose of each single ECU needs to be determined during series production.
Deep learning approaches can not easily be applied to this problem, 
because labeled training data is not available in sufficient numbers.
Thus, we train state-of-the-art artificial neural networks (ANNs) on purely synthetic training data, 
which is automatically created from a single CAD file.
By randomizing parameters during rendering of training images,
we enable inference on RGB images of a real sample part.
In contrast to classic image processing approaches, 
this data-driven approach poses only few requirements regarding the measurement setup and transfers to related use cases with little development effort.
\section{Introduction}\label{introduction}
An exemplary use case for our approach is the 3D pose estimation of electronic control units (ECUs).
The pose of each individual ECU needs to be detected robustly to enable an automated application of sealing materials.
Generally, automated image processing is widely used in industrial series production \cite{korodi_image-processing-based_2020, huang_automated_2015}.
A commonly used method to detect ECU poses is by applying classic image processing.
In some cases, these algorithms rely on fiducial marks imprinted on parts themselves.
Setting up this image processing pipeline needs to be done by experts individually for each new product.
Substituting classic image processing by fully automatically designed deep learning on our task can potentially save a significant amount of development effort for new product designs.
Furthermore, ANNs generally impose much lower requirements regarding camera resolution and surrounding conditions, enabling simpler measurement setups. 
Adding fiducial markers in industrial applications ``may be undesirable'' \cite{ren_domain_2019}.
Dropping the need for fiducial markers prevents changes on the product design, which would involve product engineers. \\
However, deep state-of-the-art ANN architectures require a large number of labeled images, which are usually not available for ECUs.
In contrast to real-world images, rendered CAD images are a widely available data source in industrial settings.
A network trained solely on CAD data cannot be directly applied to real images though, since those are different with respect to pixel color values (see \cite{tavares_de_menezes_machine_2019}).
This domain gap is generally present on many different settings that involve ANNs. \\
Techniques to overcome this domain gap are called domain adaptation and are subject to current research (e.g. \cite{bohland_influence_2019, tavares_de_menezes_machine_2019}).
Instead of adapting the ANN to a different domain, an approach can also include adaptation of the domain itself:
Images from the training domain can be randomized to such an extent, that the real-world domain is "just another variation" \cite{tobin_domain_2017}.
This method is called domain randomization and has been tested successfully on other use-cases such as indoor drone flight \cite{sadeghi_cad2rl:_2016} or robotic grasping and manipulation \cite{tobin_domain_2017, tremblay_deep_2018, akkaya_solving_2019, grun_evaluation_2019, ren_domain_2019}.
Sundermeyer et al. are working on pose estimation by using a denoising autoencoder architecture \cite{sundermeyer_implicit_2018, sundermeyer_multi-path_2020}.
In contrast to Sundermeyer et al., we are using state-of-the-art ANN architectures and evaluate different randomization parameters.
Tremblay et al., Khirodkar et al. and Hinterstoisser et al. are using domain randomization for object detection \cite{tremblay_training_2018, khirodkar_domain_2019, hinterstoisser_annotation_2019}.
Khirodkar et al. focus on the use case of detecting cars and also includes a pose estimation.
Domain randomization is not limited to image data however.
For example, Peng et al. have randomized the dynamic properties of their simulation model to transfer a robotic control algorithm trained by deep reinforcement learning to the real world \cite{peng_sim--real_2018}.
\\
In contrast to our setup, pose estimation approaches like BB8 \cite{rad_bb8:_2017}, SSD-6D \cite{kehl_ssd-6d:_2017} or PoseCNN \cite{xiang_posecnn:_2017} employ further pose refinement to improve accuracy \cite{tekin_real-time_2018, do_deep-6dpose:_2018}.
Tekin et al. \cite{tekin_real-time_2018} and Do et al. \cite{do_deep-6dpose:_2018} use standardized datasets, which does not target the domain gap that is widely present in real-world use cases.
Kleeberger et al. also use domain randomization for pose estimation, but uses depth data instead of RGB images \cite{kleeberger_single_2020}.
\\
The generally very promising results on related use cases motivate the deployment of deep ANNs for pose estimation of ECUs.
In Section \ref{chp:methods} we outline our approach in a general way.
This description is made on an abstract level without implementation details.
It serves as a template, which can easily be applied to similar use cases. \\
Subsequently, the experimental setup as described in Section \ref{chp:experimentalSetup} includes the details.
In contrast to the previous section, we set out specific implementation aspects.
A detailed overview is given e.g. over the parameters that we randomize and the way we set up the training datasets. \\
Results for our use case are presented in Section \ref{chp:results}.
Performance during inference is listed for the different training datasets.
We include an estimation of how much errors differ from their mean values. \\
In Section \ref{chp:discussion} we discuss the results. 
We analyze the effects of different randomizations of training data.
Furthermore, we compare this data-driven approach against classic image processing setups, with a focus on the possible impact on future manufacturing setups. \\
Eventually, in Section \ref{conclusion} the most relevant aspects are summarized and a detailed outlook onto further research opportunities is given.
\section{Methods}\label{chp:methods}
\begin{figure}[htb]
	\centering
	\includegraphics[width=1\textwidth]{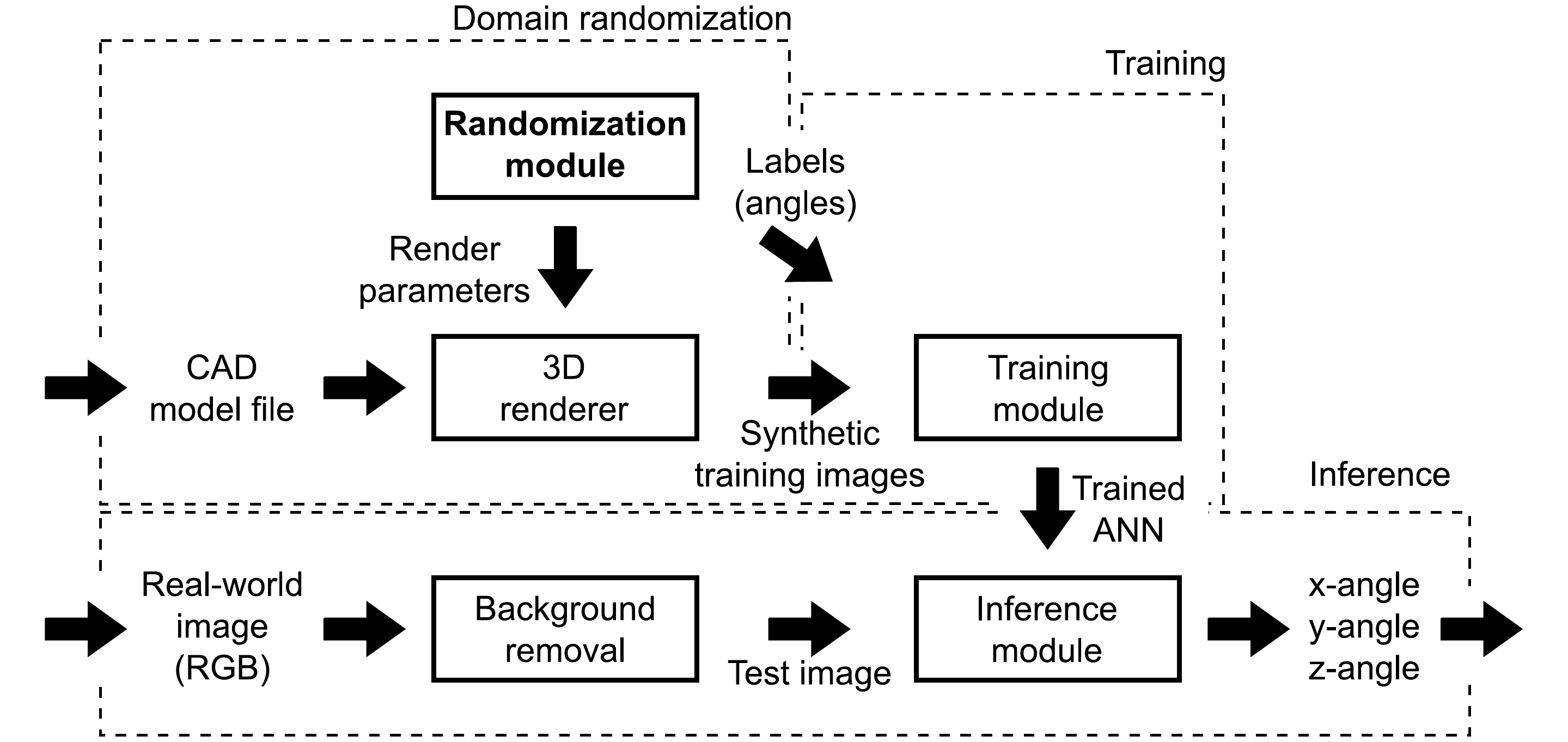}
	\caption{Overview of our approach}\label{fig:pipeline_full}
\end{figure}
Figure \ref{fig:pipeline_full} gives an overview of our approach.
To cope with the domain gap between simulated and real-world images we use domain randomization.
Unlike domain adaptation methods, domain randomization does not adapt the ANN to a different domain.
Instead, it rather adapts the training domain itself.
Our training domain consists of CAD images, which are rendered with arbitrary settings.
We generate different datasets of synthetic images automatically.
We train state-of-the-art ANNs on synthetic datasets and perform inference on real-world images.
\subsection{Domain randomization}
We created a generic pipeline for applying domain randomization as depicted in the upper left area of Figure \ref{fig:pipeline_full} (our specific implementation details are outlined in Section \ref{chp:experimentalSetup}).
CAD files are loaded into a 3D rendering software suite. 
A common exchange format is used for transferring CAD files. 
Geometric features are not changed in any way. 
Within the 3D renderer, several randomizations are applied. 
Specifically, we randomize shadows, translations and gray tones.
Those are all parameters,
which are not causally connected to the labels.
For example, gray tones have no causal connection to rotation angles and must not affect inference.
The pose of the model is set to a random angle configuration. 
An arbitrary number of angle configurations with individual randomizations is rendered and output to image files. 
Randomization parameters and rotation angles (=labels) are set via a scripting interface.
Datasets with different randomization parameters can be created in any desired number automatically. 
This kind of domain randomization is used to close the domain gap from training on CAD data to inference on real-world data (CAD2Real), saving the need for any labelled real-world images.
Once finalized, this setup can be used for different products with minimum effort by replacing the CAD model and rerunning the script.
Images for our baseline dataset \textit{CAD\_unchanged} are created with this method as well by simply omitting the randomizations.
\subsection{ANN inference on real-world images}
Real-world images are automatically preprocessed and then fed into the trained ANN.
As shown in the lower part of Figure \ref{fig:pipeline_full}, the ANN infers all three rotation angles directly from real-world RGB images.
We use a fully automatized preprocessing step to replace the background with a uniform gray tone.
\section{Experimental Setup}\label{chp:experimentalSetup}
We described our approach generically in Section \ref{chp:methods}. 
Here, we provide specific implementation details.
We outline the creation of our different datasets used for training and during inference.
Details regarding the ANN are also provided below.
\subsection{Datasets}\label{sec:datasets}
We work with two general types of datasets. 
On the one hand we have different training datasets. 
Those are solely based on CAD data and generated automatically. 
On the other hand we use experimental data for testing purposes. 
This experimental data is captured from real product samples in a laboratory setting.
\subsubsection{Training sets}
For creation of our training datasets we use the 3D rendering software \textit{Blender}.
Blender is open-source software and freely available. 
It includes a powerful Python API that enables control via automatic scripts.
The image generation pipeline is generally depicted in Figure \ref{fig:pipeline_full}.
We import the CAD model as STEP-file, a data-format commonly used for CAD data exchange.
The geometry itself is not modified.
Rotations are applied around the x-/y- and z-axis within a range of -15 to 15 degrees for the x- and y-axis and a range of -45 to 45 degrees for the z-axis.
The rotation angles serve as labels and are therefore stored for each image created.
Random translations of the model along the x- and y-axis are optionally applied within a range of -1.5\,cm and 1.5\,cm.
Random translations of the camera along the z-axis are optionally applied within a range of -1.5\,cm to 1.5\,cm.
One light source is placed high above the model, emitting uniform lighting.
Two additional light sources are optionally included as well.
Those are placed randomly above the model, generating random shadows.
The color of the model itself is randomized in uniform tones of gray.
We are aware that the introduction of additional light sources also makes the part appear brighter.
Therefore, we try to limit this effect by keeping the distance of both additional lights to the part on a constant value.
For each randomly drawn x- and y-coordinate the z-coordinate is calculated so that the distance to the part is always equal, effectively placing both lights on an imaginary sphere.
All parameters with respective ranges are listed in Table \ref{tab:parameterTable}.
\begin{table}[hbtp]
	\centering
	\caption{Parameters and their respective ranges}\label{tab:parameterTable}
	\begin{tabular}{K{35mm}  X{30mm}}
		\toprule
		\textbf{Parameter} & \textbf{Range}\\
		\toprule
		\rule{0pt}{11pt} X-/y-angle & [-15\,$^{o}$, 15\,$^{o}$]   \\
		\rule{0pt}{11pt} Z-angle & [-45\,$^{o}$, 45\,$^{o}$] \\
		\rule{0pt}{11pt} Part translations & [-1.5\,cm, 1.5\,cm] \\
		\rule{0pt}{11pt} Camera translations & [-1.5\,cm, 1.5\,cm] \\
		\rule{0pt}{11pt} Gray tones & [0.05, 0.8] \\
		\rule{0pt}{11pt}   & $x$ $\in$ [-1\,m, 1\,m] \\ 
		\rule{0pt}{11pt} Light positions & $y$ $\in$ [-1\,m, 1\,m] \\ 
		\rule{0pt}{11pt}   & z = f(x, y) \\
		\bottomrule
	\end{tabular}
\end{table}
We are using the training sets described below. \\
During evaluation we compare our domain randomization approach against two datasets:
\begin{figure}[hbtp]
	\centering
	\begin{subfigure}{0.8\textwidth}
		\centering
		\includegraphics[width=0.9\linewidth]{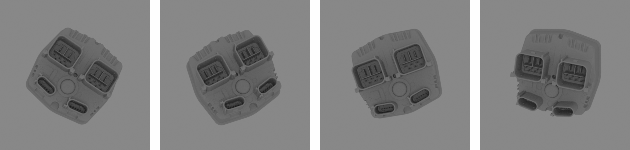}
		\caption{\textit{CAD\_unchanged}}
	\end{subfigure}
	\begin{subfigure}{0.8\textwidth}
		\centering
		\includegraphics[width=0.9\linewidth]{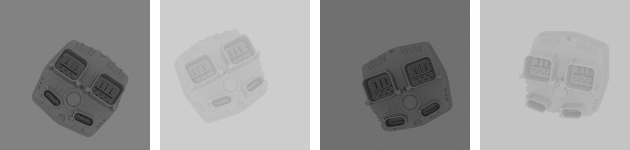}
		\caption{\textit{CAD\_augmented}}
	\end{subfigure}
	\begin{subfigure}{0.8\textwidth}
		\centering
		\includegraphics[width=0.9\linewidth]{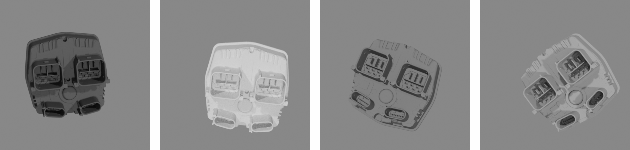}
		\caption{\textit{Rand\_full}}
	\end{subfigure}
	\begin{subfigure}{0.8\textwidth}
		\centering
		\includegraphics[width=0.9\linewidth]{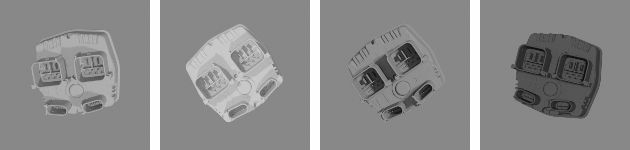}
		\caption{\textit{Rand\_noTranslations}}
	\end{subfigure}
	\begin{subfigure}{0.8\textwidth}
		\centering
		\includegraphics[width=0.9\linewidth]{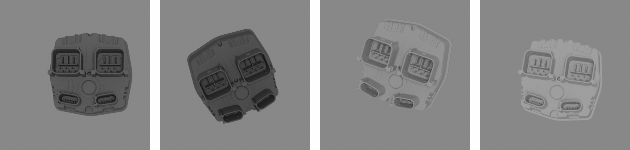}
		\caption{\textit{Rand\_noShadows}}
	\end{subfigure}
	\begin{subfigure}{0.8\textwidth}
		\centering
		\includegraphics[width=0.9\linewidth]{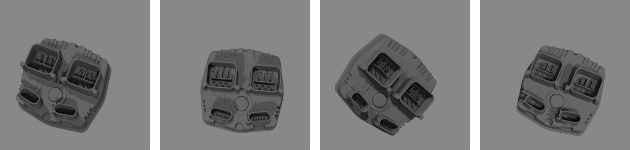}
		\caption{\textit{Rand\_noGraytones}}
	\end{subfigure}
	\caption{Example images from training datasets}\label{fig:datasets}
\end{figure}
\begin{itemize}
	\item \textit{CAD\_unchanged}: Images are labeled with rotation angles around x-/y- and z-axis. There are no further modifications made.
	\item \textit{CAD\_augmented}: We take the images from the set \textit{CAD\_unchanged} and apply random modifications. We modify brightness, translations and zoom. This data augmentation is applied to already rendered images. Data augmentation of this kind is usually used when not enough labeled training data samples are available.
\end{itemize}
The following datasets include domain randomization. To test the influence of different parameters, we modify the extent of our randomization. Samples for each dataset are shown in Figure \ref{fig:datasets}.
\begin{itemize}
	\item Dataset \textit{Rand\_full} includes all randomizations described above.
	\item Dataset \textit{Rand\_noShadows} is the same as \textit{Rand\_full} except for the two additional light sources. Thus, no random shadows are included.
	\item Dataset \textit{Rand\_noTranslations} is the same as \textit{Rand\_full} except for the translations of the part and the camera. Part and camera remain at constant positions.
	\item Dataset \textit{Rand\_noGraytones} is the same as \textit{Rand\_full} except for randomizing the uniform gray tones. Part color is only affected by the position of both randomly placed lights.
\end{itemize}
For each dataset described here, we have created 100\,000+ training images.
\subsubsection{Real-world dataset}
\begin{figure}[hbtp]
	\centering
	\includegraphics[width=0.9\textwidth]{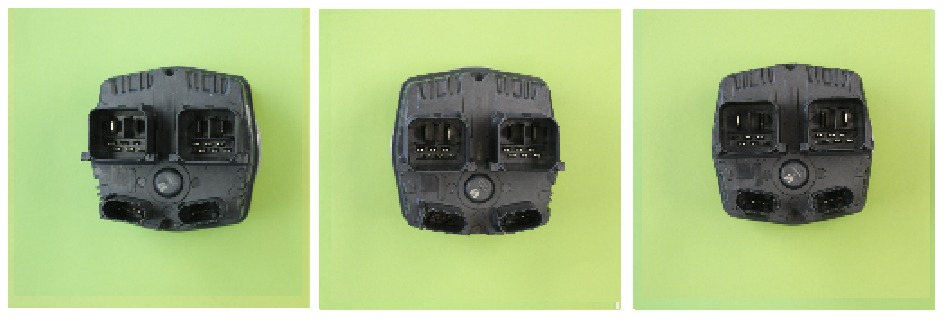}
	\caption{Example images from our real-world dataset with green background}\label{fig:realWorldDataset}
\end{figure}
We outline the creation of our real-world dataset, which is used as test set during inference.
We use a single sample part from an ECU that is already available during prototype phases (before starting series production).
We use the following laboratory setup: \\
The images are recorded with a common SLR camera, since there are no specific requirements regarding the camera.
Image resolution is later scaled down during pre-processing to below 300x300 pixels.
The camera is mounted onto a fixed frame with two lighting sources on either side.
The sample part is placed ~0.5\,m below the camera on a green cardboard layer.
To introduce rotations around the horizontal x- and y-axis we are using 3D-printed wedges.
We use multiple wedges with slopes of 2.5, 5 and 10 degrees.
Placing the wedges below our sample introduces the respective rotations.
We recorded 20 different angle configurations, examples are shown in Figure \ref{fig:realWorldDataset}.
\subsection{Artificial Neural Network: Training and inference}
We use the state-of-the-art architecture \textit{InceptionV3} \cite{szegedy_rethinking_2016}, with weights pre-trained on ImageNet.
The architecture including its weights can be imported within Keras \cite{chollet_keras_2015} in two lines of code.
To adjust to our number of labels we append a dense layer with three neurons.
Pose estimation is sometimes implemented as classification, with a binning of rotation angle intervals.
Binning limits the resolution of angle values to discrete intervals.
To avoid this, we opt for using regression as proposed by e.g. Mahendran et al. \cite{mahendran_3d_2017} instead.
We use a linear activation function within the last layer, directly outputting continuous rotation angles around the x-/y- and z-axis.
Each ANN is trained on a dataset as described above.
We train on each dataset with 10\,000 randomly drawn samples for 50 epochs.
First runs have shown slightly differing results when re-drawing the samples and re-starting training.
Therefore, we execute 30 independent runs on each dataset.
For final evaluation we use error metrics as described below.
The lower area of Figure \ref{fig:pipeline_full} shows our setup for inference with the trained ANN.
We use preprocessing of the real-world images for removing the background.
Since the images are taken on green background, preprocessing can be done automatically without much effort.
\subsection{Evaluation metrics}\label{sec:evaluationMetrics}
In this section we describe the metrics used for evaluation later on. First, we focus on the evaluation of the training set characteristics (to get an impression of ``which training set is best''). We then also estimate the confidence of statements that are made on our limited number of real-world images. \\ 
For a single test or validation image, an ANN outputs three distinct angle values $\hat{y}_{i}, i \in \{1,2,3\}$. 
These are compared against the corresponding ground truth angle values $y_{i}, i \in \{1,2,3\}$ by calculating error 
\begin{equation}\label{eq:singleError}
e_{i} = \parallel \hat{y}_{i} - y_{i} \parallel _{1}.
\end{equation}
Each single ANN is evaluated on validation and test data. 
We merge all angle values ($\hat{y}_{i}$ and $y_{i}$) into one vector per dataset ($\boldsymbol{\hat{y}}$ and $\boldsymbol{y}$).
This is done on validation and test data separately for each ANN.
Errors are then averaged over all angles and the respective number of images, yielding an mean error per ANN of 
\begin{equation}\label{eq:meanANN}
\overline{e}_{ANN} = \frac{1}{3N}  \parallel \boldsymbol{\hat{y}} - \boldsymbol{y} \parallel _{1}.
\end{equation}
$N$ is the number of images in either the validation ($N$ = 500) or test set ($N$ = 20).
Since we assess the fitness of the training set characteristics, we calculate the mean error per training set with
\begin{equation}
\overline{e} = \sum_{j=1}^{M} \overline{e}_{ANN, j},
\end{equation}
with $M$ = 30 ANNs for each training dataset.
We further calculate the standard deviation per training set
\begin{equation}
s = \sqrt{\frac{1}{M-1} \sum_{j=1}^{M} (\overline{e}_{ANN, j} - \overline{e})^{2} }.
\end{equation}
This yields a standard error of 
\begin{equation}
s_{\overline{e}} = \frac{s}{\sqrt{M}}.
\end{equation}
We further calculate a margin of error for $\overline{e}$.
Since our sample size is limited, we use the Student's t-distribution instead of the normal distribution.
For our sample size of $M$ = 30 and a confidence level of 99\% we calculate the margin of error for $\overline{e}$ as
\begin{equation}
MOE = \pm t_{M-1} s_{\overline{e}},
\end{equation}
with $t_{M-1}$ = 2.76.
When working with a normal distribution instead, $t_{M-1}$ would be replaced by $z_{c}$ with a value of 2.58 for the 99\% confidence level
($t_{M-1}$ converges towards $z_{c}$ for very large sample sizes).
The value of $t_{M-1}$ is retrieved from a table for the Student's t-distribution (see e.g. \cite{puhani_statistik:_2020}, p.206) and depends on the sample size and the confidence level.
For our sample size of $M=30$, we need a $t$-value that corresponds to \mbox{$M$ - 1 = 29} degrees of freedom.
The values of the distribution function in the table can be used directly when working with a one-sided interval.
Our two-sided interval gives an estimation into both directions (upper and lower bound).
Therefore, we record $t$ for a value of 0.995 for the distribution function to account for the two-sided interval. \\ \\
Up to now, we have mainly looked at how results differ when re-drawing samples from the training set and retraining the ANN.
The limited number of images within the test set is another factor that might affect our results.
For our limited amount of real-world samples we take a similar approach as above, but now on a more isolated scope.
For the first ANN trained on dataset \textit{Rand\_full}, we evaluate the mean error $\overline{e}_{test}$ for angles around the x-axis on the test set containing 20 images.
This is the same calculation as presented in Equation (\ref{eq:meanANN}), but discarding angles around y- and z-axis.
For those 20 error values $e_{i}$ we also calculate the standard deviation
\begin{equation}
s_{test} = \sqrt{\frac{1}{N-1} \sum_{i=1}^{N} (e_{i} - \overline{e}_{test})^{2}}
\end{equation}
and the margin of error
\begin{equation}
MOE_{test} = \pm t_{N-1} \frac{s_{test}}{\sqrt{(N)}}
\end{equation}
with $t_{N-1}$ = 2.86 for $N$ = 20 and a 99\% confidence level.
\section{Results}\label{chp:results}
We train 30 ANNs on each of the datasets described in Section \ref{sec:datasets}.
We evaluate the mean absolute error and include a margin of error as described in Section \ref{sec:evaluationMetrics} for test and validation data on each dataset. \\
Results are shown in Table \ref{tab:evaluation}.
\begin{table}[hbtp]
	\centering
	\caption{Mean error on validation data and real-world data}\label{tab:evaluation}
	\begin{tabular}{K{35mm}  X{25mm} X{25mm}}
		\toprule
		\textbf{Training dataset} & \textbf{Validation data} & \textbf{Real-world data} \\
		& $\overline{e}$ [$^{o}$]   & $\overline{e}$ [$^{o}$]  \\
		\toprule
		\rule{0pt}{11pt} \textit{CAD\_unchanged} & 0.4 $\pm$ 0.04 & 11.7 $\pm$ 2.4 \\
		\rule{0pt}{11pt} \textit{CAD\_augmented} & 0.4 $\pm$ 0.05 & 2.5 $\pm$ 0.6 \\
		\hline
		\rule{0pt}{11pt} \textit{Rand\_full} & 0.5 $\pm$ 0.06 & \textbf{1.5 $\pm$ 0.2} \\
		\rule{0pt}{11pt} \textit{Rand\_noTranslations} & 0.5 $\pm$ 0.07 & 7.7 $\pm$ 2.8 \\
		\rule{0pt}{11pt} \textit{Rand\_noShadows} & 0.5 $\pm$ 0.1 & 3.6 $\pm$ 0.8 \\
		\rule{0pt}{11pt} \textit{Rand\_noGraytones} & 0.4 $\pm$ 0.09 & \textbf{1.2 $\pm$ 0.2}  \\
		\bottomrule
	\end{tabular}
\end{table}
The outer left column indicates the dataset used during training. 
We then evaluate the performance on validation data and real-world data. 
Validation data images are from the same domain as the images used during training. 
Real-world data is taken from product samples and therefore substantially different. 
This is our target domain used for testing. \\
ANNs trained on \textit{CAD\_unchanged} exhibit the lowest error on the validation set, but insufficient performance on real-world test images. \\
For \textit{CAD\_augmented} the mean absolute error on real-world images is lower by a factor of approximately five. \\
Another improvement by another factor of almost two over \textit{CAD\_augmented} is gained by using \textit{Rand\_full}: Angles around all axes are inferred with an mean error of 1.5 degrees. \\
\textit{Rand\_full} has full randomization applied.
For \textit{Rand\_noTranslations} and \textit{Rand\_noShadows} we note an error-score inbetween \textit{CAD\_unchanged} and  \textit{CAD\_augmented}. 
Dropping the randomization of translations affects performance worse than dropping randomization of shadows. 
Training on \textit{Rand\_noGraytones} gives slightly better results than on \textit{Rand\_full}, but only by a small margin. \\
Errors on the validation set increase from \textit{CAD\_unchanged} to \textit{CAD\_augmented} and further rise for the randomized datasets. 
All randomized datasets show similar errors on validation data. \\ \\
We now take an isolated look onto the limited number of real-world samples as described in Section \ref{sec:evaluationMetrics}.
We evaluate the error for rotations around the x-axis only and look at a single ANN trained on  \textit{Rand\_full}.
For our 20 real-world samples the ANN inference has a mean error of \textbf{1.6} $\mathbf{\pm}$ \textbf{0.4 degrees}.
This margin of error is calculated for a confidence level of 99\%. \\
Our experimental setup as described in Section \ref{sec:datasets} has measurement errors which affect the ground truth labels.
All results presented above are naturally limited to measurement tolerances. \\
\section{Discussion}\label{chp:discussion}
The results presented in Section \ref{chp:results} show a consistent advantage by training on randomized datasets for our application of pose estimation of ECUs. 
In the later part of this section we also discuss the impact of this research direction on image processing setups in related product applications.
But first we look more closely at the effects of how the datasets were set up. \\
First of all, the insufficient performance with the dataset \textit{CAD\_unchanged} is not surprising. 
In this case, the training images differ a lot from the real-world images. 
This can be interpreted as a ``wide'' domain gap, leading to poor transferability from source domain to target domain. \\
A significantly improved performance on real-world images is achieved by applying state-of-the-art data augmentation to the training set.
Data augmentation is commonly used to expand the training set size.
This is especially useful when dealing with a limited amount of labeled training data.
We believe that there is a second benefit of data augmentation.
Augmenting training data with changing brightness or translations also increases the diversity within the training set.
Increased diversity of features not relevant for inference favors transferability from source to target domain.
This effect is exactly the underlying idea of domain randomization.
Data augmentation therefore can be seen as a ``light version'' of domain randomization. \\
With full domain randomization applied, inference quality is further increased by another factor of approximately two.
In comparison to data augmentation, domain randomization introduces even more diversity to the training set.
This time, the introduced diversity goes beyond simply adjusting images.
Modifications of this kind cannot be easily applied to raw images.
This is especially clear for the randomization of shadows.
Calculating the position and intensity of shadows is an integral part during rendering and not easily possible when working on two-dimensional image data only.
Also, translations made within the renderer lead to different outcomes compared to augmentation by translations as well.
Translations applied during state-of-the-art data augmentation will not change camera perspective.
In contrast, inside the renderer not only the part position changes, but also the perspective view of the part changes.
Translations in data augmentation therefore are different from those in domain randomization.
However, we believe that the major error reduction is achieved simply by the fact that additional translations are introduced, no matter whether perspective changes or not.
The poor performance with the dataset \textit{Rand\_noTranslations} especially motivates the introduction of translations. \\
To get a better impression of the different aspects of domain randomizations, 
in addition to dropping the added translations in \textit{Rand\_noTranslations}
we dropped the shadow randomization in \textit{Rand\_noShadows} and the gray tone randomization in \textit{Rand\_noGraytones}.
The performance with \textit{Rand\_noShadows} is worse compared to \textit{Rand\_full} and \textit{CAD\_augmented}.
It seems that shadows and translations are both relevant factors when dealing with the present domain gap.
However, dropping the randomization of gray tones in \textit{Rand\_noGraytones} has not caused deteriorating performance,
but even shows slight improvements compared to \textit{Rand\_full}. \\
Since we trained 30 different ANNs we do not believe this effect is caused by chance.
A possible explanation is that by randomization of gray tones many gray tones are outside of a usable scope (e.g. too light or too dark).
This leads to a smaller part of training set being useful for inference, since some images are ``too far off''.
Also, we want to mention that the effects of introducing random shadows and random gray tones overlap in some sense.
Both affect the color of the part at a certain position and are not entirely independent of each other.
In our opinion, the unexpected behavior on the dataset \textit{Rand\_noGraytones} does not hurt the idea of domain randomization in general.
It rather motivates further studies on the unique effects of different randomization types.
Instead of randomizing gray tones only, e.g. textures could be introduced as well.
\\
Also, the varying error on the validation set from CAD datasets to randomized datasets motivates the analysis of different hyperparameters, mainly training set size and the number of epochs.
The most efficient hyperparameters are generally likely to be different depending on the randomization type and extent. \\
With domain randomization we desire to cover all aspects during rendering that make the real-world images different from the CAD images.
This does not mean representing reality in simulation exactly however.
For example when dealing with differing textures of the part in the real-world, 
applying textures with random noise to the simulation might be sufficient. \\ \\
The successful application of domain randomization on this use case shows high potential for future setups of image processing pipelines in series production of ECUs.
The pipeline that we used can easily be transferred to other products.
Other products may be manufactured on different production lines.
Subsequent processing of pose information varies between products or production lines.
To standardize e.g. the naming of parameters on these interfaces and during further processing steps, an ontology-based approach is useful.
Zhou et al. \cite{zhou_predicting_2020} and Svetashova et al. \cite{svetashova_ontology-enhanced_2020} have applied ontologies to other production processes successfully.
This approach not only helps during technical setup, but also enables a common understanding of process-specific details across all involved persons \cite{zhou_predicting_2020, svetashova_ontology-enhanced_2020}. \\ \\
In contrast to many algorithms of classic image processing, our pose estimation approach is not bound to specific product features.
Further improvements aimed at improving inference accuracy can also maintain this product-independent aspect.
This advantage is based on the fact, that the features needed for inference are learned by the ANN during the training process and not manually tuned.
This data-driven approach has the advantage that for a different problem setting only the problem-specific training data needs to be supplied.
We use only data sources that are available without major effort.
With our approach, the problem-specific training data can be created automatically from a single CAD model file.
Once again we emphasize that training is done on purely synthetic data.
Not a single real-world image is needed during training.
We see a high potential for a significant reduction of development effort in future image processing setups.
\section{Summary and Outlook}\label{conclusion}
We have set out to analyze the applicability of domain randomization to our use case of pose estimation of ECUs.
Our goal was to minimize the domain gap and to deploy an ANN trained solely on synthetic data to real-world images.
We have shown that applying domain randomization exceeds the effect of data augmentation by a factor of around two.
The mean error for inferred rotations around all three axes is only 1.2 degrees on real-world images.
The entire pipeline for creating randomized training datasets and training the ANN is fully automatized.
The only input needed for creation of all training data is a single CAD model file, which is readily available for all ECUs.
We use only a state-of-the-art ANN architecture with a minor adjustment regarding the output dimensionality.
Training is done end-to-end, we infer all rotation angles directly from RGB images.
No further depth data is needed.
We have analyzed the application to our use case and motivated further research directions. \\
The following aspects could make this approach fit for application in production of ECUs:
\begin{itemize}
	\item We focused on detecting part rotations. For application in series production a post-processing step to determine translational degrees of freedom needs to be appended. Including the translations directly into the labels as well might also be a feasible approach for our use case (directly inferring 6D pose information).
	\item Our pre-processing currently requires a green-colored background. Randomizing the backgrounds as done in other use cases \cite{tobin_domain_2017, ren_domain_2019, tremblay_training_2018, sundermeyer_implicit_2018} could make our approach feasible for a background containing workpiece carriers. This would drop the need for using any pre-processing at all.
	\item We have provided insights into the effect of different randomizations. To further improve accuracy, these influences need to be examined in more detail. Ideally, this simultaneously includes adjustment of hyperparameters for training the ANN as well.
\end{itemize}
The successful execution of the steps outlined above can reduce the entire pipeline for 6D pose estimation to solely a state-of-the-art ANN architecture.
These architectures are conveniently available within the Keras library.
This would provide a fully automated pipeline for pose estimation of new ECUs and similar products.

\end{document}